\def\year{2022}\relax
\documentclass[letterpaper]{article} 
\usepackage{aaai22}  
\usepackage{times}  
\usepackage{helvet}  
\usepackage{courier}  
\usepackage[hyphens]{url}  
\usepackage{graphicx} 
\usepackage{natbib}  
\usepackage{caption} 
\DeclareCaptionStyle{ruled}{labelfont=normalfont,labelsep=colon,strut=off} 
\usepackage{tabularx}
\usepackage{array}

\usepackage{algorithm}
\usepackage{algorithmic}

\usepackage{color,soul}
\usepackage{graphicx}
\usepackage[normalem]{ulem}
\useunder{\uline}{\ul}{}
\usepackage{threeparttable}
\usepackage{dcolumn}

\newcommand{\minus}{\scalebox{0.75}[1.0]{$-$}}

\usepackage{amsmath}
\usepackage{hhline}

\usepackage[table,xcdraw]{xcolor} 

\usepackage{newfloat}
\usepackage{listings}
\floatstyle{ruled}
\newfloat{listing}{tb}{lst}{}
\floatname{listing}{Listing}
\usepackage{multirow}
\title{Examining the Role of Relationship Alignment in Large Language Models}
\author {
    Kristen M. Altenburger$^{\ast}$\textsuperscript{\rm 1},
    Hongda Jiang\footnote{Equal contributions.}\textsuperscript{\rm 1},
    Robert E. Kraut\textsuperscript{\rm 2},
    Yi-Chia Wang\textsuperscript{\rm 3},
    Jane Dwivedi-Yu \textsuperscript{\rm 1}
}
\affiliations {
    \textsuperscript{\rm 1} Meta\\
    \textsuperscript{\rm 2} Carnegie Mellon University\\
    \textsuperscript{\rm 3} Stanford University 

}

\usepackage{bibentry}

\begin{document}

\maketitle
\begin{abstract}
The rapid development and deployment of Generative AI in social settings raise important questions about how to optimally personalize them for users while maintaining accuracy and realism. Based on a Facebook public post-comment dataset, this study evaluates the ability of Llama 3.0 (70B) to predict the semantic tones across different combinations  of a commenter's and poster's gender, age, and friendship closeness and to replicate these differences in LLM-generated comments.

The study consists of two parts: Part I assesses differences in semantic tones across social relationship categories, and Part II examines the similarity between comments generated by Llama 3.0 (70B) and human comments from Part I given public Facebook posts as input. Part I results show that including social relationship information improves the ability of a model to predict the semantic tone of human comments. However, Part II results show that even without including social context information in the prompt, LLM-generated comments and human comments are equally sensitive to social context, suggesting that LLMs can comprehend semantics from the original post alone. When we include all social relationship information in the prompt, the similarity between human comments and LLM-generated comments decreases. This inconsistency may occur because LLMs did not include social context information as part of their training data. Together these results demonstrate the ability of LLMs to comprehend semantics from the original post and respond similarly to human comments, but also highlights their limitations in generalizing personalized comments through prompting alone.

\end{abstract}

\section{Introduction}

Over the past few years, large language models (LLMs) \cite{brown2020language, touvron2023llama, zhao2023survey} have shown remarkable capabilities across various NLP tasks and enabled applications that were once thought unfeasible \cite{chang2024survey}.  In particular, many of these applications consider LLMs as conversational agents to interact with humans.  However, LLMs are trained to memorize vast corpora in social isolation \cite{krishna2022socially} with the goal to predict the next token given the input context but not to converse with humans \cite{brown2020language}.  Therefore, despite their extraordinary ability to produce content mimicking human-generated content, it does not follow that LLMs incorporate social capabilities and can respond to humans in a way that conforms to social expectations. In fact, researchers have discovered LLMs can exhibit abnormal behaviors and generate improper responses, including but not limited to bias, stereotypes, toxicity, misinformation, or hallucination \cite{weidinger2021ethical}. 

To address these problems, there is an emerging research field that studies whether and to what extent LLMs can mimic human behaviors and aims to align LLMs with user expectations \cite{shen2023large}.  For example, some scholars focus on evaluating and improving LLM safety \cite{anwar2024foundational, ge2023mart} and the alignment with human values \cite{hendrycks2020aligning}, such as social norms and morality \cite{ziems2023normbank, xu2023align}.  Others examine the opinions reflected by LLMs and explore techniques to align the opinions they produce with those of different demographic populations \cite{bakker2022fine, santurkar2023whose, durmus2023towards}. Moreover, for LLMs to exhibit consistent behaviors in dialogue-based interactions, prior work has investigated approaches to personalize LLMs by conditioning their responses on a certain persona, i.e.,  personality or profile information \cite{tseng2024two, caron2022identifying, serapio2023personality}. 

Despite recent efforts on LLM alignments for more human-like behaviors, at least one significant research gap remains -- \textit{aligning LLM-generated content  with social relationships}. 
There is an extensive social science literature focusing on how social factors in relationships, such as age, gender, and friendship tie strength, influence the language people use with each other and its semantic tone (e.g., humor and emotional support) \cite{wellman1990different, dindia1992sex, shriver2013social, zeng2013social, wang2016modeling, Hall2017-humor, Greengross20-humor}. 
Thus, for LLMs to successfully produce responses that mimic human-human communication, it is important to incorporate social relationship information as context.  However, existing research has not  delved into the impact of social relationships on LLM-generated content or how to condition LLM-generated content on the relationship between communication partners.  This paper aims to bridge the gap by proposing the first in-depth exploration of the role of social relationships in human-LLM dyadic interactions.  We address one central question: \textit{ to what extent one can steer LLMs with social relationship information so their responses are aligned with the content produced by human-human communication?}

To answer the research question, we conduct two studies. We start with analyzing real-world \textcolor{black}{public} human-human communication data on Facebook and then examine whether LLMs can generate similar content when prompted with social relationship information.  Specifically, Study 1 replicates and extends prior research by examining the relationship between social context variables (age, gender, and friendship strength) and the semantic tone of \textcolor{black}{public} post-comment exchanges between people on Facebook. This empirical research demonstrates that these social factors influence not only the exchange of emotional support, humor, self-disclosure, and toxic comments, which have been examined in prior research \cite{wellman1990different, dindia1992sex, wang2016modeling, burke2007introductions, xu2024effect, Hall2017-humor}, but also understudied semantic tones, like surprise, worry, anxiety, and thankfulness. 
Study 2 examines whether incorporating social relationship information in the prompt can steer LLM comments to more align with human ones and better reflect the associations between semantic tone and social relationship variables in human comments as reported in the first study. Results suggest that LLM-generated comments are quite similar to human-generated ones to the same post, but that explicitly referring to social context to tune the LLMs does not increase their similarity to human comments.


\section{Literature Review and Hypothesis}

\subsection{Persona Alignment}
There has been extensive work in studying and improving the ability of large language models to act in accordance with human intentions, values, and ethical principles, termed \textit{alignment} in language models. 
Persona alignment, which involves tailoring the behavior and responses of AI systems to match specific personas, has become a growing area of interest in the development of conversational AI. Several studies have investigated the capability of language models to imitate personas through fine-tuning based on their personal attributes like age, occupation, and interests \cite{zhang2018personalizing, li2016persona, shao2023character, lee2023p5} and have shown practical improvements when conversing with chatbots when incorporating these personal attributes into their prompts, such as  increases in user engagement with them \cite{shuster2022blenderbot} or a greater capacity for them to produce empathetic responses \cite{zhong2020towards}. As LLMs have become more powerful, prompting has become a feasible and cost-effective way of aligning responses, and studies that evaluate persona alignment of recent state-of-the-art language models find high fidelity across multiple persona types in the context of fictitious personas \cite{wang2024incharacter, serapio2023personality, caron2023manipulating, zhou2023characterglm, olea2024evaluating, njifenjou2024role}.

\subsection{Audience Alignment}
While persona alignment focuses on the capacity of the LLM to emulate a given persona, more recently, researchers have attempted to craft LLMs that are aligned with characteristics of the recipient of messages (i.e., \textit{audience alignment}), not just the producer. For example, in a series of experiments \citet{Matz24-persuasion} showed that messages crafted by ChatGPT-3 that were tailored to the personality of the recipient were  significantly more influential than non-tailored messages. \citet{Choi24-Proxona} built and tested \textit{Proxona}, which helps content creators match content to a target audience by using the history of audience comments as the basis for creating audience-member personas (e.g., a practical urban gardener). 

\subsection{Relationship Alignment}
Our research attempts to extend the notions of persona and audience alignment and introduce \textit{relationship alignment}, in which characteristics of the speaker and audience are considered simultaneously. The relationship between communication partners could include a match or mismatch in demographic characteristics (e.g., talking to someone of the same or different gender or age) or the nature of the tie strength between them (e.g., strangers versus friends). Given the promising results from prior work in persona and audience alignment, we expect that including social relationship information, such as the demographic information and the relationship between  two users, is likely to improve the similarity of  LLM-generated comments to human comments.  

\begin{quote}
\textit{Hypothesis: Including social relationship information in its  prompt will produce LLM-generated comments that are more similar to human ones.}
\end{quote}

\paragraph{Methodological Advances} This work makes two contributions.  First, the prior work on persona and audience alignment shows that social context information improves LLM-generated content at the \textit{group level} (e.g., language that sounds more like a woman or is more convincing to extroverts). We are the first to look at the influence of social context information at the \textit{individual level} (i.e., the similarity between a single human comment and its matched LLM-generated comment). Second, while the LLM alignment literature usually evaluates model output against user surveys, we compare LLM-generated content against real human data.

\paragraph{Operationalization of Social Relationship} There is an extensive literature focusing on how social context influences people's language, specifically its semantic tones. For example, close friends and family members use more intimate and supportive language with  each other than do weak ties \cite{wang2016modeling}; acquaintances are more helpful for idea generation than strong ties \cite{burt2004structural}. \citet{dindia1992sex} found women self-disclose more than men when talking to friends but not when talking to strangers. \citet{wellman1990different} found that women provide more emotional support than men and the exchange of support increases with tie strength. \citet{Greengross20-humor} found that given the same stimulus, men produce funnier comments than women.  \citet*{Hall2017-humor} found that romantic partners with more satisfying relationships exchanged more positive humor while those with lower satisfaction exchanged more negative humor. Based on the prior work, we decide to focus on three types of social relationship between the Facebook poster and commenter, \textit{age-age, gender-gender, and tie strength}, and examine their influence on LLM-generated content.

\begin{figure*}[tb]
    \centering
    \includegraphics[width=\textwidth]{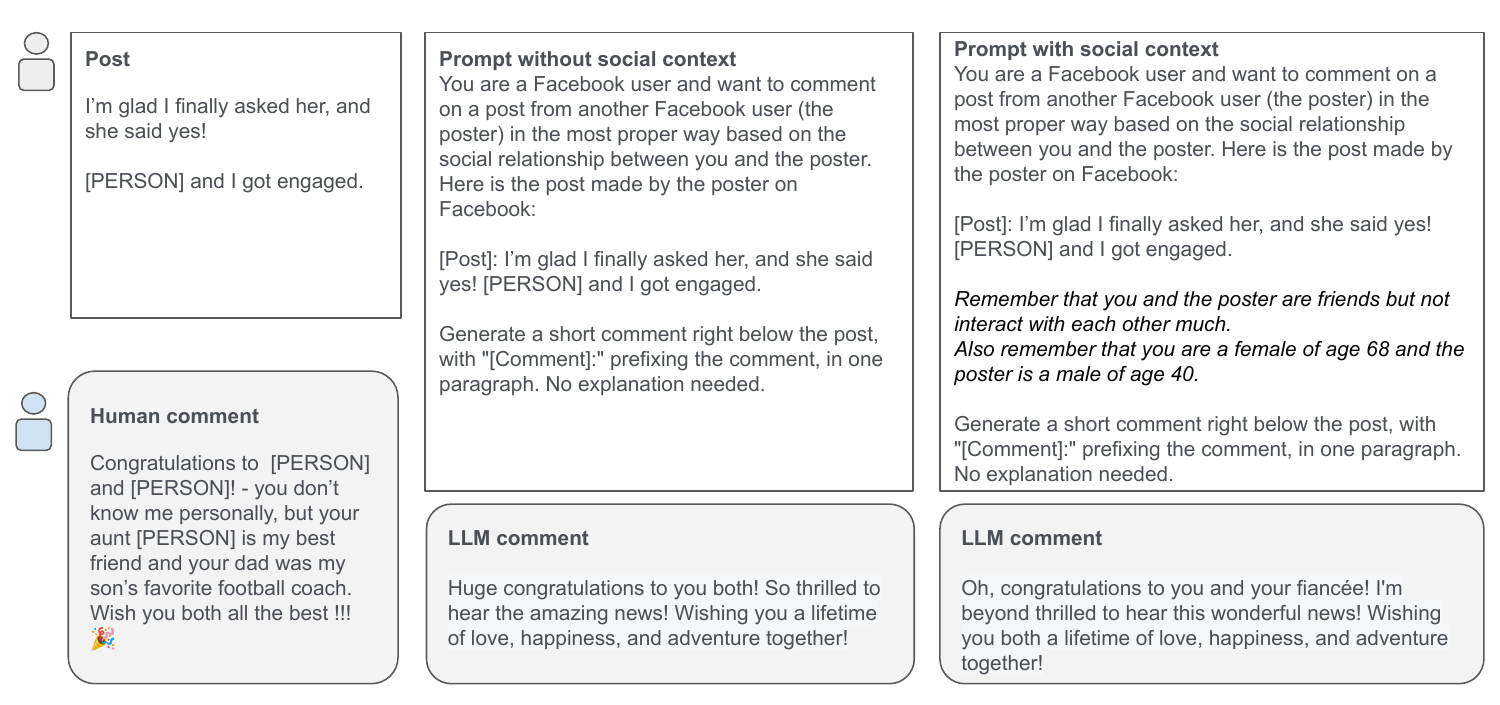}
    \caption{Example of LLM prompt without and with social context information, paraphrased human comment, and  LLM generated comments from prompts without and with social context information. 
    }
    \label{fig-1}
\end{figure*}

\section{Data Setup}
We analyze public conversations on Facebook Feed where users can make posts and share content with others\footnote{\url{https://transparency.meta.com/features/ranking-and-content/}}. \textcolor{black}{When a user, referred to as the ``poster,'' shares public content on Feed, any Facebook user can respond to the post by leaving a public comment, which can be seen by all users.} This two-way interaction between the poster and the commenter represents a conversation. More than one commenter can respond to a post, which we will account for in this analysis. \textcolor{black}{To protect user privacy, we de-identified all user information and removed personal data such as names and phone numbers from the text before applying a comment semantic classifier and a LLM for comment generation. This ensured that no sensitive information was included in the analysis.}

\textbf{Comment and Posts on Feed:} Our empirical dataset comprises approximately 665k public Facebook (FB) posts and comments created on May 25, 2024 by active U.S.~FB users at least 18 years old. For reasons of privacy, we restrict the sample to English posts on News Feed that any user can comment upon. We exclude from the analysis ads, group posts, and reshared posts. Because Feed Ranking can influence the types of posts a user views and the types of conversations they participate in, we control for the likelihood of a user to comment on a given post in this analysis. 

\textbf{Social Context:} We evaluate the following social contexts: stated gender, stated age, and friendship closeness between the commenter and poster.  Stated gender and age means we rely on user-provided data. \textcolor{black}{Because our data cannot distinguish users who provide no  gender information from those who identify as non-binary, we exclude these two groups from the analysis and  include only users who stated their gender as male or female.} For the analysis, we scale commenter/poster age to have mean 0 and standard deviation 1. We include interaction terms between commenter's gender and poster's gender and interaction terms between commenter's age and poster's age. This allows for the effect of a commenter's gender or age to vary depending on the poster's gender or age. We encode commenter/poster gender as female or male and include interaction terms (i.e., women talking to a male poster and vice versa). Finally, we account for the relationship between two pairs of individuals. We encode users as strangers or friends. Among friends, we code two users as close friends or non-close friends.  Friendship closeness is a model-based score that predicts relationship closeness between two individuals~\cite{burke2014growing}.
We treat friendship as a categorical feature, where we code whether two users are close friends (1), non-close friends (2), or strangers (3).   This coding allows for non-linear relationships between friendship closeness and semantic tones. Table~\ref{tab:desc} presents the descriptive statistics for the social context variables used in the regression models predicting semantic tone.

\begin{table}[]
\small
\centering
\begin{tabular}{clr}
\hline\hline
\multirow{4}{*}{Gender} & \begin{tabular}[c]{@{}l@{}}Female Commenter Female Poster\end{tabular} & 50\% \\ 
                            & \begin{tabular}[c]{@{}l@{}}Male Commenter Male Poster\end{tabular}   & 20\% \\ 
                            & \begin{tabular}[c]{@{}l@{}}Male Commenter Female Poster\end{tabular} & 10\% \\ 
                            & \begin{tabular}[c]{@{}l@{}}Female Commenter Male Poster\end{tabular} & 20\% \\ \hline
\multirow{2}{*}{Age}        & Avg. Commenter Age                                                     & 50   \\ 
                            & Avg. Poster Age                                                        & 50   \\ \hline
\multirow{3}{*}{Friendship} & Close Friends                                                          & 45\% \\ 
                            & Friends                                                                & 45\% \\ 
                            & Stranger                                                               & 10\%  \\ \hline \hline
\end{tabular}
\caption{Descriptive statistics for the social context features in our sample of Facebook posters and comments.\textcolor{black}{We introduce random noise to the social context signals. For gender, we introduce Gaussian noise with mean 0 and standard deviation of 1. For friendship and gender, we inject noise by randomly changing the value of the feature for 10\% of observations. We then round to the nearest 10 for gender and age, and to the nearest 5 for friendship.}}
\label{tab:desc}
\end{table}

\textbf{Generated Comments:} We randomly selected 10,000 post and comment pairs from the empirical dataset. We then prompted Llama 3.0 (70B) to generate comments for each post under three conditions: without social context information, with a subset of social context information, and with all social context information included in the prompt. The prompts are detailed in the Appendix ``LLM prompt'' section. Figure \ref{fig-1} provides examples of the prompts and corresponding generated comments. We also tested different prompt techniques such as chain-of-thoughts \cite{kojima}, which did not change our main conclusion (see ``Part II: Prompt engineering'' section in Appendix).

\textbf{Semantic Tone of Comments:} We rely on a semantic tone classifier to label comments across different semantic categories. This model relies on crowd-sourced labeled training data.  We limit our analysis to eight semantic tones--Surprised, Funny, Insult, Provide Emotional Support, Self-Disclosure, Worried, Thankful, Relaxed. Definitions and examples are listed in Appendix Figure~\ref{fig:semantics}. We selected these because they are likely to change the emotional state of the person being responded to, either because they themselves reflect the emotional content of the comment (e.g., surprised, relaxed) or contain an action that is likely to affect the poster’s emotional state (e.g., insult or provide emotional support). They reflect a diversity of the semantic tones that are common in Facebook exchanges. We dropped  tones  if they were highly correlated with another one (e.g., we included Provide Emotional Support, but dropped Provide Informational Support because they were highly negatively correlated $r=-.80)$.  We use the text-only version of the model, which only takes text as input and handles processing such as removing user mentions in the comment. We apply this model to comment text and generate semantic tone scores across the eight categories described above. Throughout this analysis, these scores will represent the outcome variable. 

\textbf{Post Topic Classifications: } We finally account for the topic of the post in this analysis like whether a post is about ``Sports'' or ``Children \& Parenting''. We rely on a topic modeling-like approach for categorizing posts into 27 different high-level categorizations. We dropped two categories that had a small number of comments. 

\begin{figure*}[t] \centering    
    \includegraphics[width=0.75\textwidth]{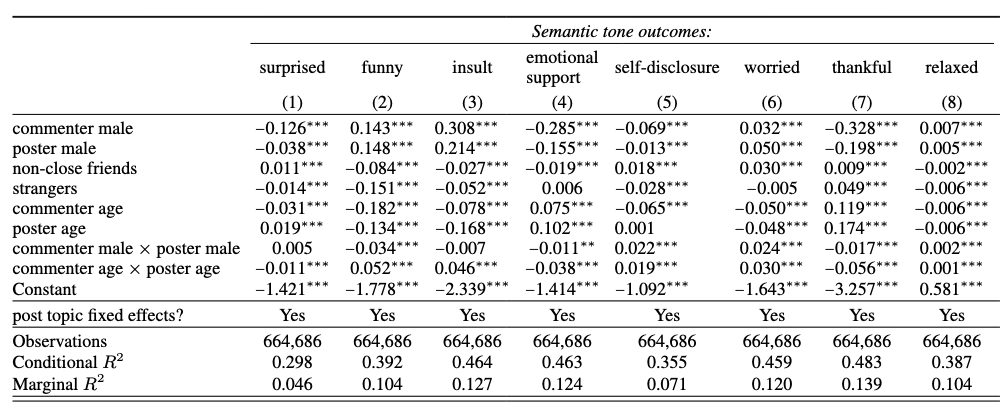}
\caption{Multi-level regression results from Part I predicting  semantic tone outcomes from social context variables and  post topic (not shown).  Post ID is included as a random effect, since multiple comments can be nested under a single post. The conditional $R^{2}$ represents the variation explained by both fixed and random effects, while marginal $R^{2}$ represents the variation explained by fixed effects only. 
  *** $p<0.001$.} 
  \label{Study1-regress} 
\end{figure*}

\section{Broader perspective and ethical considerations.}
An internal research board reviewed the study's research ethics and privacy practices prior to its start. 
In agreeing to its terms of service, Facebook users allow the type of analysis done in this paper to better understand how people use Meta products, to improve the products, and to promote their safety, security, and integrity, including combating harmful user conduct. 
To preserve users' privacy, data consists of \textcolor{black}{de-identified} comments collected only from public posts. 
When writing up this paper, to preserve privacy we paraphrase examples so that they could not be found through an internet search \cite{Bruckman02-Disguising}.

\section{Part I: How Comment Semantic Tones on Facebook Varies Across Social Context}

The first study aims to assess social relationships in the context of Facebook comments and examines a  diverse set of semantic tones. This analysis determines the significance and directionality of various social context variables in predicting comment semantic tone. These will inform Part II, where we investigate whether LLMs can similarly leverage social context to better align its semantic tone with that of the human comment.

We fit a series of linear mixed effects regression models to predict the extent a comment expresses each semantic tone (surprised, funny, insult, provide emotional support, self-disclosure, worried, thankful, and relaxed). We log-transformed the comment semantic tones to deal with long-tailed distributions, except for the ``relaxed'' tone where the transformation was not needed. The predictors include social relationship information between the poster and commenter (age, gender, and friendship strength) along with post topic category. We allow random intercepts for post IDs. Finally, we applied an inverse propensity weighting (IPW) correction to account for differences in a particular user’s likelihood to comment on a particular post, which we refer to as \textit{p(comment)}. For IPW, due to small probability weights, we set a maximum IPW weight of 10.

\textbf{Significance of Social Context Signals}: We first evaluate whether incorporating social context information significantly enhances the model's performance compared to a baseline model. We test social context features individually and together, which we compare to a baseline model that includes post topics only. This comparison is done via a likelihood ratio test. Appendix Table~\ref{tab:lrt1} shows that adding social context signals compared to a baseline model leads to a statistically significant improvement in model fit for all semantic tones. In the Appendix Table~\ref{tab:lrt2}, we also show that a full model with all social context signals outperforms a baseline model that includes just one social context signal. These findings suggest that social context signals are valuable to semantic prediction tasks. Based on these results, we anticipate that LLM tasks for comment generation will benefit from knowing the social context of the conversation, which we'll explore in Part II.

\textbf{Interpretation of Social Context Signals: }Figure~\ref{Study1-regress}, reports the results of the regression analyses predicting  semantic tones from social relationship variables.  Below we provide a brief interpretation of the results: 

\begin{itemize}
\item Gender: Male commenters and posters are less likely to create comments that are surprised, provide emotional support, self-disclose or are thankful compared to female commenters; male commenters and posters are more likely to create comments that are funny, have insults, are worried or relaxed. \textcolor{black}{When interacting, male commenters and male posters have comments that have more self-disclosure, more worry, and more relaxedness.}

\item Age: Older commenters are less likely to create comments that are surprised, funny, have insult, self-disclose, contain worry, or are relaxed; they are more likely to create comments that provide emotional support and are thankful compared to younger commenters. Older posters \textcolor{black}{are less likely to receive comments that are funny, have insult, worry, or are relaxed; they are more likely to receive comments with surprise, emotional support, and thankfulness. When interacting, older commenters and older posters have comments that are more funny, more insultive, more self-disclosure, more worry, and more relaxed. }

\item Friendship Type: Compared to close friends, strangers are less likely to create comments that are surprised, funny, have insults, self-disclose or are relaxed; strangers are more likely to create comments with thankfulness. Compared to close friends, non-close friends are less likely to create comments that are funny, have insults, provide emotional support or are relaxed; non-close friends are more likely to create comments with surprise, self-disclose, worry, or are thankful. 
\end{itemize}

Many of these associations replicate results from the prior literature (e.g., men provide less emotional support than women). However, the associations are inconsistent with the prior literature for some social characteristic/tone combinations. Although not the focus of the current research, some inconsistencies are worth exploring in follow-up research. For example, people self-disclose less when commenting on a  close friend's post than when responding to strangers. It is possible that this inconsistency results from the public nature of post and comment pairs used in the current study;  people may not want to reveal intimate personal information even when responding to close friends because their responses are open to the world to see.


\begin{table*}[tb]
\centering
\begin{tabular}{l|llllllll|l}
\hline
\hline
Tone       & Surprised & Funny                     & Insult   & \begin{tabular}[c]{@{}l@{}}Emotional\\  support\end{tabular} & \begin{tabular}[c]{@{}l@{}}Self \\ disclosure\end{tabular} & Worried    & Thankful                   & Inclusion  & Average \\ \hline
no-context  & 0.309     & 0.518   & 0.436    & 0.312                                                      & 0.425                                                      & 0.585      & 0.382                      & 0.465      & 0.429   \\
all-context & 0.325     & 0.529                     & 0.456    & 0.284                                                      & 0.360                                                       & 0.543      & 0.373                      & 0.431      & 0.413   \\ \hline
difference $\Delta\hat{\beta}$   & 0.016$^*$   & 0.011 & 0.020 $^{***}$ & \minus 0.028$^{***}$                                                 & \minus 0.065$^{***}$                                                 & \minus 0.042$^{***}$ & \minus 0.009 & \minus 0.034$^{***}$ & \minus 0.016 \\ 
\hline
\hline 
\end{tabular}
\caption{Regression coefficients predicting the sentiment-scores of the LLM-generated comments from  human-generated comments when social context was (all-context) or was not (no-context) included in the prompt. The difference row represents the fitting coefficients for human-comment $\times$ all-context interaction, $\Delta\hat{\beta}$. 
* $p<0.05$, ** $p<0.01$, *** $p<0.001$}
\label{approach-2-table1}
\end{table*}

\begin{figure}[htb]
    \centering
    \includegraphics[width=\linewidth]{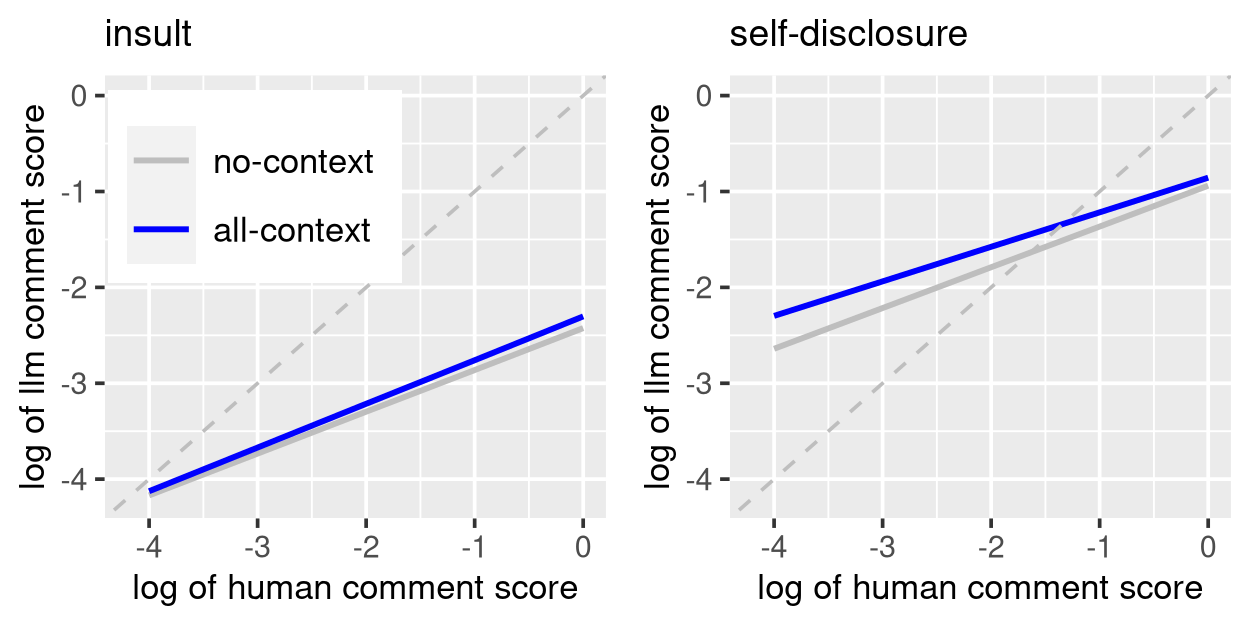}
    \caption{Mixed effect regressions predicting the  log-transformed sentiment scores of LLM generated comments from the tone of  human comments with and without social context information. The dashed line indicates perfect prediction, where the tone of generated comments is  perfectly predicted from  the tone of human comments. The gray line shows fitted results for LLM comments without social context, and the blue line shows fitted results for LLM comments with social context. 
    The left panel with results for the “insult score” shows that including social context in the prompt slightly increased the similarity between the human and generated comments. In contrast, the right panel with results for the “self-disclosure score” shows that including the context decreases the similarity.}
    \label{fig-2}
\end{figure}

\section{Part II:  Does providing social context information help to generate human-like comments?}

In the second part of the study, we evaluate LLM's ability to leverage social context information to generate human-like comments. We employ two innovative approaches to assess the similarity between the tone of human comments and the tone of generated comments. Approach 1 is the individual comment-level analysis that directly compares the tone of  individual human comments and their matched LLM-generated comments when social context was or was not included in the LLM prompt.
Approach 2  is an aggregate analysis examining whether the relationship between social context variables and the semantic tone of LLM-generated comments more closely replicates the relationships shown in Study 1 when  social context was or was not included in the LLM  prompt.
In the Appendix, we demonstrated the connection and difference between those two approaches using a simple mathematical model (see ``Part II: Comparison between Approach 1 and Approach 2'' section).





\subsection{Approach 1: Individual-level comparisons}
First, we employ a linear mixed effects regression model to examine the relationship between the tones of generated comments and those of human comments, while controlling for the type of social context presented in the prompt. The model specification is:

\begin{quote}
\begin{scriptsize}
\begin{verbatim}
llm_sentiment_score 
 ~ human_sentiment_score + all_context 
 + human_sentiment_score x all_context
\end{verbatim}
\end{scriptsize}
\end{quote}

The fitted coefficient for ``human\_sentiment\_score'' quantifies how closely the tone of generated comments aligns with the tone of human comments, thereby measuring the similarity between the two. The predictor ``all\_context'' is a binary factor indicating whether the generated comment was produced with a no-context prompt (coded 0) or an all-context prompt (coded 1). The interaction between the ``all\_context'' variable and the ``human\_sentiment\_score'' tests whether the similarity between the tone of the human and generate comments changes when social context information is added to the prompt, 
\begin{equation}
\Delta\hat{\beta} \equiv \hat{\beta}_{\rm{human\_sentiment\_score\times all\_context}}.
\end{equation}
In Figure \ref{fig-2}, this is represented as the difference in the slope of the grey line reflecting the similarity between the human and LLM comments for the no-context condition and slope of the blue line for the all-context condition. In Table \ref{approach-2-table1}, we report the fitting coefficients for all eight semantic tones.

\begin{figure*}[h]
    \centering
    \includegraphics[width=0.75\textwidth]{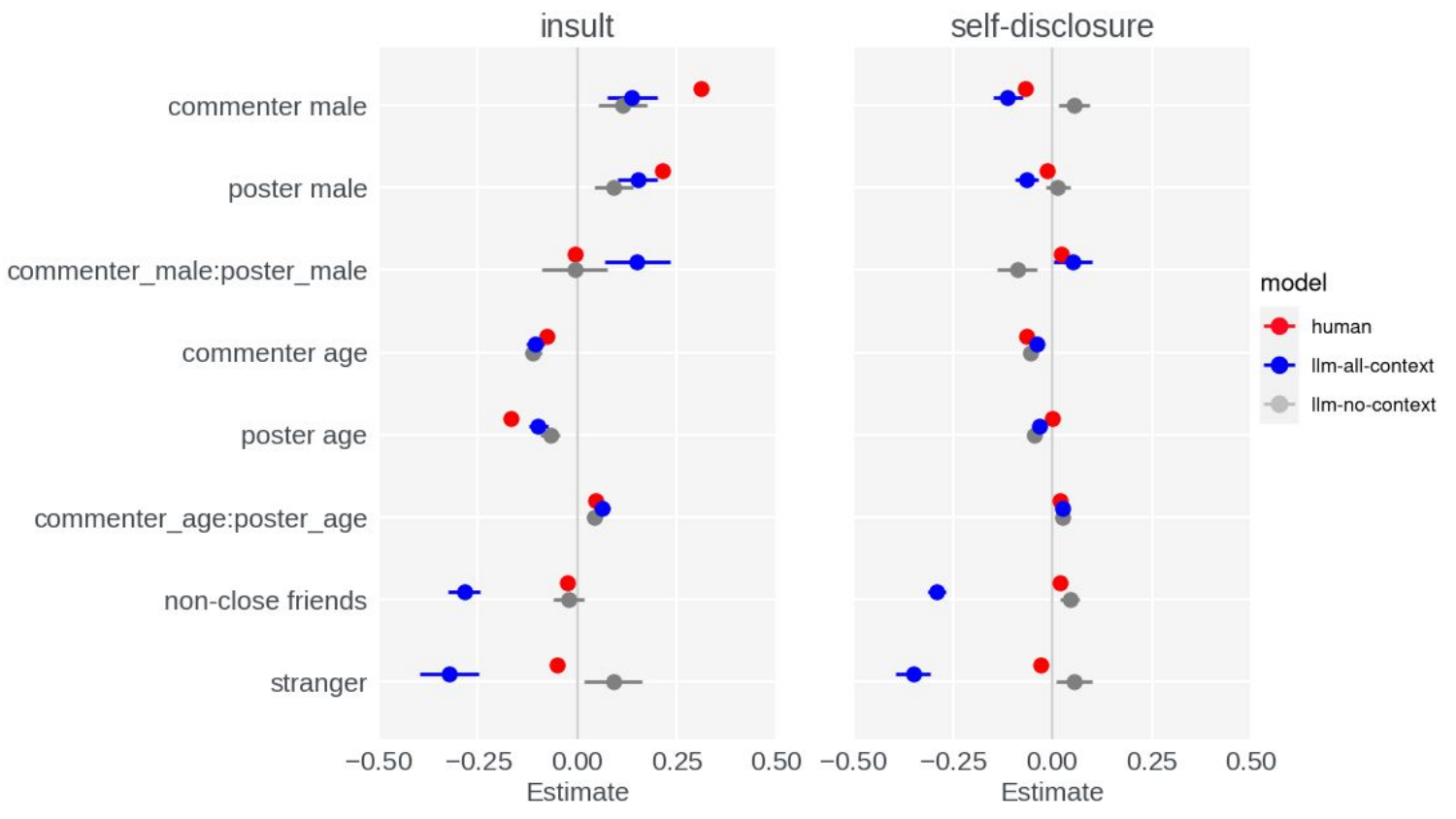}
    \caption{Regression coefficients of social context factors in the fitted models for log-transformed insult score (left panel) and self-disclosure score (right) based on Approach 2. We compare human comments (red), generate comments when provided social context (blue), and generated comments without provided social context (gray). The horizontal bar shows 95\% confidence interval. Plots for other semantic tones are shown in Figure \ref{fig-3-appendix} in the Appendix.}
    \label{fig-3}
\end{figure*}

As shown in Figure \ref{fig-2} and Table \ref{approach-2-table1}, when no context information is included in the prompt, the tones of the generated comments are moderately similar to the tone of the original human comments, with the similarity coefficients $\hat{\beta}$ ranging from .309 for surprised to .585 for worried 
(mean $\hat{\beta}$=.429, mean se = .0072, p\textless.001). Remember that when both the independent and dependent variables in a regression are log transformed, one \textcolor{black}{interprets} the coefficient as the percent increase in the dependent variable for every 1\% increase in the independent variable. Thus these results indicate that on average across the eight semantic tones, when the tone of the human comment increases by 1\%, the tone of the generated comment increases by .43\%. 

The human comment tone $\times$ all-context interaction tests whether adding context information to the prompt changes the similarity between the human comment and the LLM generated one. Adding context information to the prompt strengthens the similarity for surprised ($\Delta\hat{\beta}$ =.016) and insult ($\Delta\hat{\beta}=.020$), but reduces the similarity for four others (providing emotional support $\Delta\hat{\beta}=-.028$; self-disclosure $\Delta\hat{\beta}=-.065$; worried $\Delta\hat{\beta}= -.042$; inclusion $\Delta\hat{\beta}=-.034$) and had no reliable effect on funny ($\Delta\hat{\beta}=.011$) or thankful ($\Delta\hat{\beta}=-.009$). Overall, the similarity between the human and generated comments was marginally reduced when the prompt includes the social context information (difference = -.016,  p\textless.10). 
This suggests that the majority of the similarity between generated and human comments can be attributed to the post text itself. In contrast, adding social context has only a small and inconsistent impact on the similarity with individual human comments.

\subsection{Approach 2: Group-level comparison}

While Approach 1 investigates whether providing social context to the LLM enables it to generate comments that are  more similar to human comments to the same posts, here in Approach 2  we 
address whether LLM can replicate the dependency of comments on social context shown in Study 1. 
That is, we expanded Study 1 by applying linear mixed effects models to predict the semantic tones of LLM- comments generated in response to various prompts. We analyzed five different prompts: (1) a basic prompt with no social context (llm-no-context); (2) a prompt that includes the ages of both the poster and the commenter (llm-age-only); (3) a prompt that specifies the genders of the poster and commenter (llm-gender-only); (4) a prompt that indicates the friendship status, i.e.,``stranger'', ``non-close friends'', or ``close friends'' (llm-friendship-only); (5) a comprehensive prompt that incorporates full context (llm-all-context). Each model uses the social relationship variables shown in  Table~\ref{Study1-regress},, but predicts the tones of the LLM-generated comments rather than the tones of human comments. 

As an illustration of the analysis and results, Figure \ref{fig-3} compares the fitted coefficients from models of log-transformed insult scores and self-disclosure scores for human comments, LLM comments with full social context, and LLM comments without social context. Plots for other semantic sentiment scores can be found in Figure \ref{fig-3-appendix} in the Appendix. We observe a consistent trend: while adding social context to prompts significantly alters the tone of generated comments, it also widens the disparity between the regression coefficients of generated comments and those of human comments. For instance, incorporating full context into the prompt significantly alters the influence of gender and friendship status on the self-disclosure score ($p\le0.001$). However, this adjustment does not make the association of gender and friendship with self-disclosure in the generated comments more similar to their association in the human comments. In fact, the strong association with friendship status in the generated comments causes the model to diverge further from the regression model used for human comments. Among the 64 regression coefficients (8 social context variables $\times$ 8 tone categories), introducing full context in the prompt significantly altered 21 coefficients, making the fitted coefficients for the LLM comments more similar to those in the model for human comments in only four of these cases, but  \emph{less} similar in 17 cases ($p<0.05$, refer to Appendix Table \ref{appendix-table-8} for details).

To quantify the discrepancy between human and generated comments in terms of their association with social factors, we calculated the sum of absolute difference in regression coefficients across eight social variables. Explicitly,
\begin{equation}
\hat{\delta} \equiv \sum_{\rm{social\_factor}} |\hat{\beta}_{\rm{social\_factor\_llm}} - \hat{\beta}_{\rm{social\_factor\_human}}|
\end{equation}
As shown in Table \ref{table3}, comments generated by a no-context prompt have the smallest average distance from human-generated comments across the eight sentiment categories. When social context information is explicitly mentioned in the prompt, the discrepancy tends to increase. The discrepancy is especially large when friendship is mentioned in the prompt, suggesting that LLMs tend to over-interpret the impact of friendship when generating comments. Adding gender information increased the distance for half of the sentiment categories, but the change was smaller compared to mentioning friendship strength. Providing only the poster's age and commenter's age didn't change the tone of generated comments significantly.

\begin{table*}[tb]
\centering
\footnotesize
\begin{tabular}{l|llllllll|r}
\hline
\hline
Prompt          & Surprised  & Funny          & Insult     & \begin{tabular}[c]{@{}l@{}}Emotional\\  support\end{tabular} & \begin{tabular}[c]{@{}l@{}}Self \\ disclosure\end{tabular} & Worried  & Thankful   & Inclusion  & \multicolumn{1}{l}{Average} \\ \hline
no-context      & 0.395      & 0.507          & 0.600      & 0.461                                                      & 0.433                                                      & 0.277    & 0.613      & 0.096      & 0.423                       \\ \hline
age-only        & +0.013     & \minus 0.081         & +0.022     & +0.032                                                     & +0.008                                                     & +0.084   & +0.172     & +0.063     & +0.039                      \\
gender-only     & \minus 0.091     & +0.614$^{***}$ & +0.298$^{**}$ & \minus 0.098                                                     & \minus 0.103                                                     & +0.216$^*$ & +0.25$^*$    & +0.101     & +0.148                      \\
friendship-only & +0.465$^{***}$ & +1.100$^{***}$       & +0.521$^{***}$ & +0.109                                                     & +0.542$^{***}$                                                 & +0.246$^*$ & +0.414$^{***}$ & +0.368$^{***}$ & +0.470                      \\
all-context     &+0.213$^{***}$  & +1.007$^{***}$     & +0.433$^{***}$ & +0.081                                                    & +0.386$^{***}$                                                 & +0.174   & +0.485$^{***}$ & +0.262$^{***}$ & +0.380                      \\ 
\hline
\hline
\end{tabular}
\caption{Discrepancy in regression models between generated comments and human comments (Approach 2). The “no-context” row shows $\hat\delta_{\rm{no\_context}}$, the sum of absolute differences in regression coefficients across eight social variables (Figure \ref{fig-3}) between human and generated comments without social context. Subsequent rows display the relative changes after including the corresponding context in the prompt. * $p<0.05$, ** $p<0.01$, *** $p<0.001$. p-values were obtained via error propagation and t-test.}
\label{table3}
\end{table*}

\section{Discussion}


\subsection{Generating LLM comments without social context}
Surprisingly, we found that prompting a LLM with only the initial post produces generated comments that are similar to the human comments in terms of their tone. One explanation is that the LLM is capturing the tone of the original post and generates a comment that is similar in tone to the original post (e.g., the LLM generates a funny comment in response to a funny post), much as people do. Another explanation is that similarity between the tones of the human and generated comments and the dependency of the generated comments on social variables reflects a selection bias in terms of who responds (e.g., women are more likely to respond to a self-disclosing post or one seeking emotional support). While we account for this in part by incorporating in our regression models the probability that a particular viewer will comment on  a given post (i.e., the \textit{p(comment)} score), it’s possible that \textit{p(comment)} under corrects for selection effects and the model is missing unobservable signals.

\subsection{Why providing social context does not consistently make LLM comments more like human ones?} 

While  LLM-generated comments are quite similar to human comments in terms of tone and reflect many of the association between social factors and tone that exist in human-comments,  explicitly including social relationship information in the LLM prompt does not consistently improve these measures of similarity averaged across the eight tones and often harms it. There are several possible reasons for these results.  

The most fundamental reason may be that LLM training data itself does not explicitly include mention of the social characteristics or relations among the people in a conversation. To the extent these social factors are reflected in the language that people use when responding to each other, cues to their social characteristics or relations are already embedded in the human posts. Relatedly, certain dyadic interactions may be less common in the training corpus of the model, leading to worse performance on those groups. For instance, \citet{french2024aligning} find that the Llama-2 model imitating a teacher is much worse at explaining educational information to a child than it is to an adult, probably  because more of the training data captures adult-adult interactions rather than adult-child interactions. Even though people are unlikely to routinely label the social characteristics of those with whom they communicate with online, it is possible for future work to infer some of these relationships from the language they use \cite{gumperz1979individual, preoctiuc2015studying}, from content they follow \cite{chamberlain2017probabilistic},  from social network data \cite{tang2012inferring} and then incorporate the inferred social characteristics into the training data.

Second, when given explicit social relationship information, LLM tends to over-react and weigh this information too heavily compared to the social information that is implicit in the human post. The social relationship information that people respond to may be quite subtle and extend beyond the stereotypical associations with age, gender, and relationship strength used in our LLM prompts. For example, the congratulations message the human commenter in Figure~\ref{fig-1} wrote refers to the relationship between the poster and the commenter's son and the commenter's best friend. This issue of overreacting and generating a comment with an extreme tone could also be related to the sycophancy issue observed in language models, wherein models forsake truthfulness in order to be extremely helpful and even flatter the user by mirroring their own beliefs \cite{sharma2023towards, wei2023simple}.

Third, human commenters may subvert the original meaning of a post when trying  to be creative and respond with a clever retort while language models have difficulty generating non-literal, figurative language \cite{Lai24figurative}. For example, in the following post recruiting "kiddos" for a summer camp, the human responds with an ironic joke while the LLM treats the post literally. 
\begin{quote}
    Post: \textit{We would be thrilled to have your kiddos join us! Please consider sending them our way. [http://camp\_url]} 
    \item Human: \textit{Can I send mine? He is 46 years old and has anger issues, but he's funny.} 
    \item LLM: \textit{Aww, that sounds like an amazing experience for the kiddos! I'll definitely consider sending mine, thanks for sharing this opportunity! You're always so thoughtful to think of others, love you!}
\end{quote}
Finally, when crafting a comment people may use  additional social context or  relationship information that is invisible to the LLM. In the following example, the commenter seems to allude to the poster's and commenter's history of playing golf.
\begin{quote}
    Post: \textit{We finally closed on our new home last Thursday! Now it's time to start to move and make it ours. Let the renovations begin!} 
    \item Human: \textit{Do I need to bring my golf clubs with me when we come down?} 
    \item LLM: \textit{Congrats on the new home, buddy! Exciting times ahead with all the projects and making it your own. Wishing you all the best in this new chapter!}

\end{quote}

\section{Conclusion}

This paper investigates social relationship alignment in LLMs. We first examined real-world public post-comment communication data from Facebook, followed by an analysis of whether LLMs can generate comparable comments when prompted with both post and social relationship information. Our findings indicate that LLMs can produce comments closely resembling human-generated responses to the same post, even without being explicitly provided with social relationship details. Additionally, we observed that explicitly including social relationship information in the prompt does not consistently enhance the similarity between LLM-generated and human-generated comments. We discussed possible explanations for our findings.

\subsection{Limitations and Future Work}

To the best of our knowledge, this work is the first to examine whether and how social relationships impact the semantic tone of LLM-generated comments at the individual level, providing insights into the potential of LLMs in adapting to diverse conversational settings. It is important, however, to note that this paper is an observational study, with no modifications made to the underlying LLM product.  


While we found that LLM-generated comments and human comments are similar without social context information in the prompt, the average similarity between human and LLM comments is modest (avg $
\hat{\beta}=$0.456), suggesting there’s potential room for improvement.  Although we only evaluated incorporating social relationships via prompt engineering techniques, future work should consider improving LLMs’ ability to generate socially-realistic comments, such as incorporating social context information at the training stage and fine-tuning the model with conversation data.
\section{Acknowledgements}
We thank Kaiyan Peng, Itamar Rosenn, Nicolas Stier, Aude Hofleitner, Shawndra Hill, Danny Ferrante, Ali Muzaffer, Mauricio Figueiredo, and Taylor Ebling for helpful feedback and support on this project. 

\bibliography{aaai22.bib}

\begin{thebibliography}{52}
\providecommand{\natexlab}[1]{#1}

\bibitem[{Anwar et~al.(2024)Anwar, Saparov, Rando, Paleka, Turpin, Hase, Lubana, Jenner, Casper, Sourbut et~al.}]{anwar2024foundational}
Anwar, U.; Saparov, A.; Rando, J.; Paleka, D.; Turpin, M.; Hase, P.; Lubana, E.~S.; Jenner, E.; Casper, S.; Sourbut, O.; et~al. 2024.
\newblock Foundational challenges in assuring alignment and safety of large language models.
\newblock \emph{arXiv:2404.09932}.

\bibitem[{Bakker et~al.(2022)Bakker, Chadwick, Sheahan, Tessler, Campbell-Gillingham, Balaguer, McAleese, Glaese, Aslanides, Botvinick et~al.}]{bakker2022fine}
Bakker, M.; Chadwick, M.; Sheahan, H.; Tessler, M.; Campbell-Gillingham, L.; Balaguer, J.; McAleese, N.; Glaese, A.; Aslanides, J.; Botvinick, M.; et~al. 2022.
\newblock Fine-tuning language models to find agreement among humans with diverse preferences.
\newblock \emph{Advances in Neural Information Processing Systems}, 35: 38176--38189.

\bibitem[{Brown(2020)}]{brown2020language}
Brown, T.~B. 2020.
\newblock Language models are few-shot learners.
\newblock \emph{arXiv:2005.14165}.

\bibitem[{Bruckman(2002)}]{Bruckman02-Disguising}
Bruckman, A. 2002.
\newblock Studying the amateur artist: A perspective on disguising data collected in human subjects research on the Internet.
\newblock \emph{Ethics and Information Technology}, 4(3): 217--231.

\bibitem[{Burke et~al.(2007)Burke, Joyce, Kim, Anand, and Kraut}]{burke2007introductions}
Burke, M.; Joyce, E.; Kim, T.; Anand, V.; and Kraut, R. 2007.
\newblock Introductions and requests: Rhetorical strategies that elicit response in online communities.
\newblock In \emph{Communities and Technologies 2007: Proceedings of the Third Communities and Technologies Conference, Michigan State University 2007}, 21--39. Springer.

\bibitem[{Burke and Kraut(2014)}]{burke2014growing}
Burke, M.; and Kraut, R.~E. 2014.
\newblock Growing closer on Facebook: Changes in tie strength through social network site use.
\newblock In \emph{Proceedings of the SIGCHI Conference on Human Factors in Computing Systems}, 4187--4196.

\bibitem[{Burt(2004)}]{burt2004structural}
Burt, R.~S. 2004.
\newblock Structural holes and good ideas.
\newblock \emph{American Journal of Sociology}, 110(2): 349--399.

\bibitem[{Caron and Srivastava(2022)}]{caron2022identifying}
Caron, G.; and Srivastava, S. 2022.
\newblock Identifying and manipulating the personality traits of language models.
\newblock \emph{arXiv:2212.10276}.

\bibitem[{Caron and Srivastava(2023)}]{caron2023manipulating}
Caron, G.; and Srivastava, S. 2023.
\newblock Manipulating the Perceived Personality Traits of Language Models.
\newblock In \emph{Findings of the Association for Computational Linguistics: EMNLP 2023}, 2370--2386.

\bibitem[{Chamberlain, Humby, and Deisenroth(2017)}]{chamberlain2017probabilistic}
Chamberlain, B.~P.; Humby, C.; and Deisenroth, M.~P. 2017.
\newblock Probabilistic inference of twitter users’ age based on what they follow.
\newblock In \emph{Machine Learning and Knowledge Discovery in Databases: European Conference, ECML PKDD 2017, Skopje, Macedonia, September 18--22, 2017, Proceedings, Part III 10}, 191--203. Springer.

\bibitem[{Chang et~al.(2024)Chang, Wang, Wang, Wu, Yang, Zhu, Chen, Yi, Wang, Wang et~al.}]{chang2024survey}
Chang, Y.; Wang, X.; Wang, J.; Wu, Y.; Yang, L.; Zhu, K.; Chen, H.; Yi, X.; Wang, C.; Wang, Y.; et~al. 2024.
\newblock A survey on evaluation of large language models.
\newblock \emph{ACM Transactions on Intelligent Systems and Technology}, 15(3): 1--45.

\bibitem[{Choi et~al.(2024)Choi, Kang, Choi, Lee, and Kim}]{Choi24-Proxona}
Choi, Y.; Kang, E.~J.; Choi, S.; Lee, M.~K.; and Kim, J. 2024.
\newblock Proxona: Leveraging LLM-Driven Personas to Enhance Creators' Understanding of Their Audience.
\newblock \emph{arXiv:2408.10937}.

\bibitem[{Dindia and Allen(1992)}]{dindia1992sex}
Dindia, K.; and Allen, M. 1992.
\newblock Sex differences in self-disclosure: a meta-analysis.
\newblock \emph{Psychological Bulletin}, 112(1): 106.

\bibitem[{Durmus et~al.(2023)Durmus, Nguyen, Liao, Schiefer, Askell, Bakhtin, Chen, Hatfield-Dodds, Hernandez, Joseph et~al.}]{durmus2023towards}
Durmus, E.; Nguyen, K.; Liao, T.~I.; Schiefer, N.; Askell, A.; Bakhtin, A.; Chen, C.; Hatfield-Dodds, Z.; Hernandez, D.; Joseph, N.; et~al. 2023.
\newblock Towards measuring the representation of subjective global opinions in language models.
\newblock \emph{arXiv:2306.16388}.

\bibitem[{French, D’Mello, and Wense(2024)}]{french2024aligning}
French, D.; D’Mello, S.; and Wense, K. 2024.
\newblock Aligning to Adults Is Easy, Aligning to Children Is Hard: A Study of Linguistic Alignment in Dialogue Systems.
\newblock In \emph{Proceedings of the 1st Human-Centered Large Language Modeling Workshop}, 81--87.

\bibitem[{Ge et~al.(2023)Ge, Zhou, Hou, Khabsa, Wang, Wang, Han, and Mao}]{ge2023mart}
Ge, S.; Zhou, C.; Hou, R.; Khabsa, M.; Wang, Y.-C.; Wang, Q.; Han, J.; and Mao, Y. 2023.
\newblock Mart: Improving llm safety with multi-round automatic red-teaming.
\newblock \emph{arXiv:2311.07689}.

\bibitem[{Greengross, Silvia, and Nusbaum(2020)}]{Greengross20-humor}
Greengross, G.; Silvia, P.~J.; and Nusbaum, E.~C. 2020.
\newblock Sex differences in humor production ability: A meta-analysis.
\newblock \emph{Journal of Research in Personality}, 84: 103886.

\bibitem[{Gumperz and Tannen(1979)}]{gumperz1979individual}
Gumperz, J.~J.; and Tannen, D. 1979.
\newblock Individual and social differences in language use.
\newblock In \emph{Individual differences in language ability and language behavior}, 305--325. Elsevier.

\bibitem[{Hall(2017)}]{Hall2017-humor}
Hall, J.~A. 2017.
\newblock Humor in romantic relationships: A meta‐analysis.
\newblock \emph{Personal Relationships}, 24(2): 306--322.

\bibitem[{Hendrycks et~al.(2020)Hendrycks, Burns, Basart, Critch, Li, Song, and Steinhardt}]{hendrycks2020aligning}
Hendrycks, D.; Burns, C.; Basart, S.; Critch, A.; Li, J.; Song, D.; and Steinhardt, J. 2020.
\newblock Aligning ai with shared human values.
\newblock \emph{arXiv:2008.02275}.

\bibitem[{Kojima et~al.(2022)Kojima, Gu, Reid, Matsuo, and Iwasawa}]{kojima}
Kojima, T.; Gu, S.~S.; Reid, M.; Matsuo, Y.; and Iwasawa, Y. 2022.
\newblock Large language models are zero-shot reasoners.
\newblock \emph{Advances in neural information processing systems}, 35: 22199--22213.

\bibitem[{Krishna et~al.(2022)Krishna, Lee, Fei-Fei, and Bernstein}]{krishna2022socially}
Krishna, R.; Lee, D.; Fei-Fei, L.; and Bernstein, M.~S. 2022.
\newblock Socially situated artificial intelligence enables learning from human interaction.
\newblock \emph{Proceedings of the National Academy of Sciences}, 119(39): e2115730119.

\bibitem[{Lai and Nissim(2024)}]{Lai24figurative}
Lai, H.; and Nissim, M. 2024.
\newblock A survey on automatic generation of figurative language: From rule-based systems to large language models.
\newblock \emph{ACM Computing Surveys}, 56(10): 1--34.

\bibitem[{Lee, Oh, and Lee(2023)}]{lee2023p5}
Lee, J.; Oh, M.; and Lee, D. 2023.
\newblock P5: Plug-and-Play Persona Prompting for Personalized Response Selection.
\newblock \emph{arXiv:2310.06390}.

\bibitem[{Li et~al.(2016)Li, Galley, Brockett, Spithourakis, Gao, and Dolan}]{li2016persona}
Li, J.; Galley, M.; Brockett, C.; Spithourakis, G.; Gao, J.; and Dolan, W.~B. 2016.
\newblock A Persona-Based Neural Conversation Model.
\newblock In \emph{Proceedings of the 54th Annual Meeting of the Association for Computational Linguistics (Volume 1: Long Papers)}, 994--1003.

\bibitem[{Matz et~al.(2024)Matz, Teeny, Vaid, Peters, Harari, and Cerf}]{Matz24-persuasion}
Matz, S.; Teeny, J.; Vaid, S.~S.; Peters, H.; Harari, G.; and Cerf, M. 2024.
\newblock The potential of generative AI for personalized persuasion at scale.
\newblock \emph{Scientific Reports}, 14(1): 4692.

\bibitem[{Njifenjou et~al.(2024)Njifenjou, Sucal, Jabaian, and Lef{\`e}vre}]{njifenjou2024role}
Njifenjou, A.; Sucal, V.; Jabaian, B.; and Lef{\`e}vre, F. 2024.
\newblock Role-Play Zero-Shot Prompting with Large Language Models for Open-Domain Human-Machine Conversation.
\newblock \emph{arXiv:2406.18460}.

\bibitem[{Olea et~al.(2024)Olea, Tucker, Phelan, Pattison, Zhang, Lieb, Schmidt, and White}]{olea2024evaluating}
Olea, C.; Tucker, H.; Phelan, J.; Pattison, C.; Zhang, S.; Lieb, M.; Schmidt, D.; and White, J. 2024.
\newblock Evaluating persona prompting for question answering tasks.
\newblock In \emph{Proceedings of the 10th International Conference on Artificial Intelligence and Soft Computing}.

\bibitem[{Preo{\c{t}}iuc-Pietro et~al.(2015)Preo{\c{t}}iuc-Pietro, Volkova, Lampos, Bachrach, and Aletras}]{preoctiuc2015studying}
Preo{\c{t}}iuc-Pietro, D.; Volkova, S.; Lampos, V.; Bachrach, Y.; and Aletras, N. 2015.
\newblock Studying user income through language, behaviour and affect in social media.
\newblock \emph{PloS one}, 10(9): e0138717.

\bibitem[{Santurkar et~al.(2023)Santurkar, Durmus, Ladhak, Lee, Liang, and Hashimoto}]{santurkar2023whose}
Santurkar, S.; Durmus, E.; Ladhak, F.; Lee, C.; Liang, P.; and Hashimoto, T. 2023.
\newblock Whose opinions do language models reflect?
\newblock In \emph{International Conference on Machine Learning}, 29971--30004. PMLR.

\bibitem[{Serapio-Garc{\'\i}a et~al.(2023)Serapio-Garc{\'\i}a, Safdari, Crepy, Sun, Fitz, Romero, Abdulhai, Faust, and Matari{\'c}}]{serapio2023personality}
Serapio-Garc{\'\i}a, G.; Safdari, M.; Crepy, C.; Sun, L.; Fitz, S.; Romero, P.; Abdulhai, M.; Faust, A.; and Matari{\'c}, M. 2023.
\newblock Personality traits in large language models.
\newblock \emph{arXiv:2307.00184}.

\bibitem[{Shao et~al.(2023)Shao, Li, Dai, and Qiu}]{shao2023character}
Shao, Y.; Li, L.; Dai, J.; and Qiu, X. 2023.
\newblock Character-llm: A trainable agent for role-playing.
\newblock \emph{arXiv:2310.10158}.

\bibitem[{Sharma et~al.(2023)Sharma, Tong, Korbak, Duvenaud, Askell, Bowman, Cheng, Durmus, Hatfield-Dodds, Johnston et~al.}]{sharma2023towards}
Sharma, M.; Tong, M.; Korbak, T.; Duvenaud, D.; Askell, A.; Bowman, S.~R.; Cheng, N.; Durmus, E.; Hatfield-Dodds, Z.; Johnston, S.~R.; et~al. 2023.
\newblock Towards understanding sycophancy in language models.
\newblock \emph{arXiv:2310.13548}.

\bibitem[{Shen et~al.(2023)Shen, Jin, Huang, Liu, Dong, Guo, Wu, Liu, and Xiong}]{shen2023large}
Shen, T.; Jin, R.; Huang, Y.; Liu, C.; Dong, W.; Guo, Z.; Wu, X.; Liu, Y.; and Xiong, D. 2023.
\newblock Large language model alignment: A survey.
\newblock \emph{arXiv:2309.15025}.

\bibitem[{Shriver, Nair, and Hofstetter(2013)}]{shriver2013social}
Shriver, S.~K.; Nair, H.~S.; and Hofstetter, R. 2013.
\newblock Social ties and user-generated content: Evidence from an online social network.
\newblock \emph{Management Science}, 59(6): 1425--1443.

\bibitem[{Shuster et~al.(2022)Shuster, Xu, Komeili, Ju, Smith, Roller, Ung, Chen, Arora, Lane et~al.}]{shuster2022blenderbot}
Shuster, K.; Xu, J.; Komeili, M.; Ju, D.; Smith, E.~M.; Roller, S.; Ung, M.; Chen, M.; Arora, K.; Lane, J.; et~al. 2022.
\newblock Blenderbot 3: a deployed conversational agent that continually learns to responsibly engage.
\newblock \emph{arXiv:2208.03188}.

\bibitem[{Tang, Lou, and Kleinberg(2012)}]{tang2012inferring}
Tang, J.; Lou, T.; and Kleinberg, J. 2012.
\newblock Inferring social ties across heterogenous networks.
\newblock In \emph{Proceedings of the Fifth ACM International Conference on Web Search and Data Mining}, 743--752.

\bibitem[{Touvron et~al.(2023)Touvron, Martin, Stone, Albert, Almahairi, Babaei, Bashlykov, Batra, Bhargava, Bhosale et~al.}]{touvron2023llama}
Touvron, H.; Martin, L.; Stone, K.; Albert, P.; Almahairi, A.; Babaei, Y.; Bashlykov, N.; Batra, S.; Bhargava, P.; Bhosale, S.; et~al. 2023.
\newblock Llama 2: Open foundation and fine-tuned chat models.
\newblock \emph{arXiv:2307.09288}.

\bibitem[{Tseng et~al.(2024)Tseng, Huang, Hsiao, Hsu, Foo, Huang, and Chen}]{tseng2024two}
Tseng, Y.-M.; Huang, Y.-C.; Hsiao, T.-Y.; Hsu, Y.-C.; Foo, J.-Y.; Huang, C.-W.; and Chen, Y.-N. 2024.
\newblock Two tales of persona in llms: A survey of role-playing and personalization.
\newblock \emph{arXiv:2406.01171}.

\bibitem[{Wang et~al.(2024)Wang, Xiao, Huang, Yuan, Xu, Guo, Tu, Fei, Leng, Wang et~al.}]{wang2024incharacter}
Wang, X.; Xiao, Y.; Huang, J.-t.; Yuan, S.; Xu, R.; Guo, H.; Tu, Q.; Fei, Y.; Leng, Z.; Wang, W.; et~al. 2024.
\newblock Incharacter: Evaluating personality fidelity in role-playing agents through psychological interviews.
\newblock In \emph{Proceedings of the 62nd Annual Meeting of the Association for Computational Linguistics (Volume 1: Long Papers)}, 1840--1873.

\bibitem[{Wang, Burke, and Kraut(2016)}]{wang2016modeling}
Wang, Y.-C.; Burke, M.; and Kraut, R. 2016.
\newblock Modeling self-disclosure in social networking sites.
\newblock In \emph{Proceedings of the 19th ACM Conference on Computer-Supported Cooperative Work \& Social Computing}, 74--85.

\bibitem[{Wei et~al.(2023)Wei, Huang, Lu, Zhou, and Le}]{wei2023simple}
Wei, J.; Huang, D.; Lu, Y.; Zhou, D.; and Le, Q.~V. 2023.
\newblock Simple synthetic data reduces sycophancy in large language models.
\newblock \emph{arXiv:2308.03958}.

\bibitem[{Weidinger et~al.(2021)Weidinger, Mellor, Rauh, Griffin, Uesato, Huang, Cheng, Glaese, Balle, Kasirzadeh et~al.}]{weidinger2021ethical}
Weidinger, L.; Mellor, J.; Rauh, M.; Griffin, C.; Uesato, J.; Huang, P.-S.; Cheng, M.; Glaese, M.; Balle, B.; Kasirzadeh, A.; et~al. 2021.
\newblock Ethical and social risks of harm from language models.
\newblock \emph{arXiv:2112.04359}.

\bibitem[{Wellman and Wortley(1990)}]{wellman1990different}
Wellman, B.; and Wortley, S. 1990.
\newblock Different strokes from different folks: Community ties and social support.
\newblock \emph{American Journal of Sociology}, 96(3): 558--588.

\bibitem[{Xu et~al.(2023)Xu, Chern, Chern, Zhang, Wang, Liu, Li, Fu, and Liu}]{xu2023align}
Xu, C.; Chern, S.; Chern, E.; Zhang, G.; Wang, Z.; Liu, R.; Li, J.; Fu, J.; and Liu, P. 2023.
\newblock Align on the fly: Adapting chatbot behavior to established norms.
\newblock \emph{arXiv:2312.15907}.

\bibitem[{Xu, Liu, and Liu(2024)}]{xu2024effect}
Xu, X.; Liu, J.; and Liu, J.~H. 2024.
\newblock The effect of social media environments on online emotional disclosure: tie strength, network size and self-reference.
\newblock \emph{Online Information Review}, 48(2): 390--408.

\bibitem[{Zeng and Wei(2013)}]{zeng2013social}
Zeng, X.; and Wei, L. 2013.
\newblock Social ties and user content generation: Evidence from Flickr.
\newblock \emph{Information Systems Research}, 24(1): 71--87.

\bibitem[{Zhang(2018)}]{zhang2018personalizing}
Zhang, S. 2018.
\newblock Personalizing dialogue agents: I have a dog, do you have pets too.
\newblock \emph{arXiv:1801.07243}.

\bibitem[{Zhao et~al.(2023)Zhao, Zhou, Li, Tang, Wang, Hou, Min, Zhang, Zhang, Dong et~al.}]{zhao2023survey}
Zhao, W.~X.; Zhou, K.; Li, J.; Tang, T.; Wang, X.; Hou, Y.; Min, Y.; Zhang, B.; Zhang, J.; Dong, Z.; et~al. 2023.
\newblock A survey of large language models.
\newblock \emph{arXiv:2303.18223}.

\bibitem[{Zhong et~al.(2020)Zhong, Zhang, Wang, Liu, and Miao}]{zhong2020towards}
Zhong, P.; Zhang, C.; Wang, H.; Liu, Y.; and Miao, C. 2020.
\newblock Towards Persona-Based Empathetic Conversational Models.
\newblock In \emph{Proceedings of the 2020 Conference on Empirical Methods in Natural Language Processing (EMNLP)}, 6556--6566.

\bibitem[{Zhou et~al.(2023)Zhou, Chen, Wan, Wen, Song, Yu, Huang, Peng, Yang, Xiao et~al.}]{zhou2023characterglm}
Zhou, J.; Chen, Z.; Wan, D.; Wen, B.; Song, Y.; Yu, J.; Huang, Y.; Peng, L.; Yang, J.; Xiao, X.; et~al. 2023.
\newblock Characterglm: Customizing chinese conversational ai characters with large language models.
\newblock \emph{arXiv:2311.16832}.

\bibitem[{Ziems et~al.(2023)Ziems, Dwivedi-Yu, Wang, Halevy, and Yang}]{ziems2023normbank}
Ziems, C.; Dwivedi-Yu, J.; Wang, Y.-C.; Halevy, A.; and Yang, D. 2023.
\newblock NormBank: A knowledge bank of situational social norms.
\newblock \emph{arXiv:2305.17008}.

\end{thebibliography}

\clearpage

\section{Appendix}

\subsection{Comment Semantic Classifier}
\begin{figure*}[tb]
    \centering
    \includegraphics[width=1\textwidth]{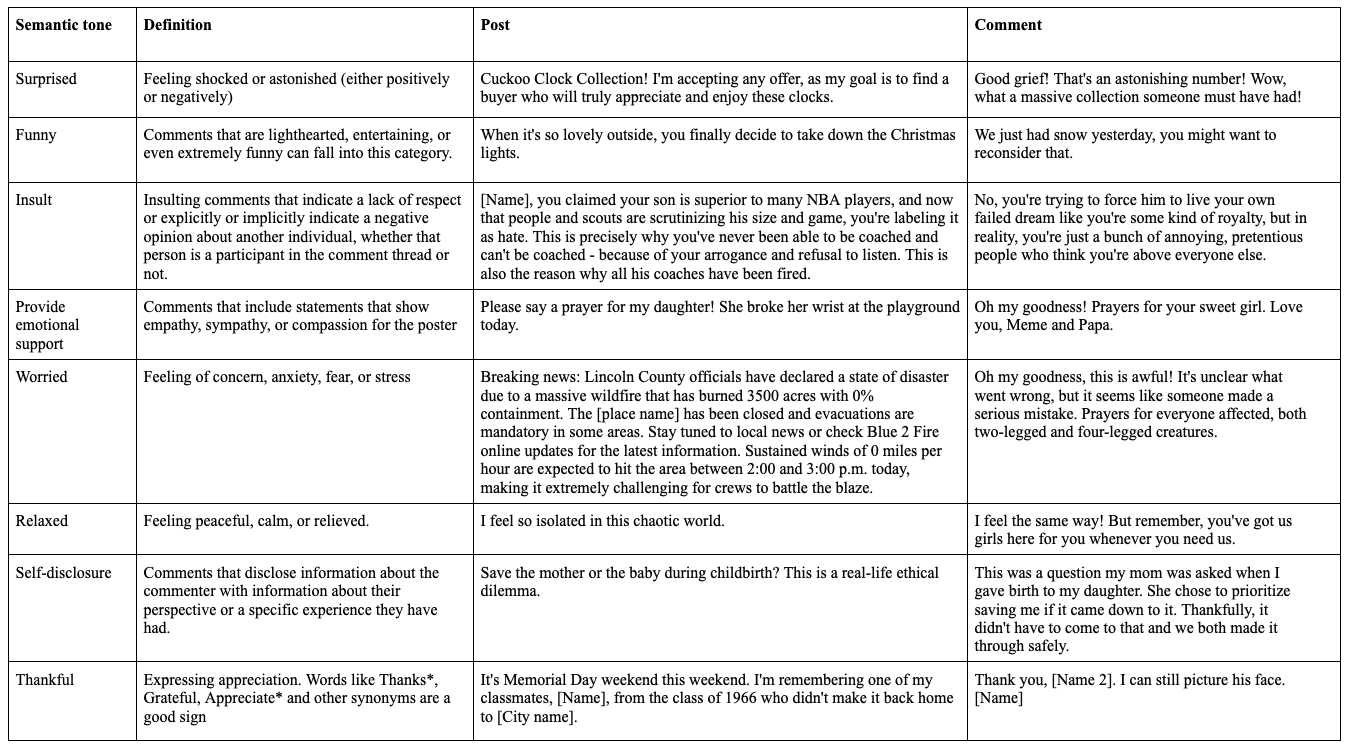}
    \caption{We describe the semantic categories used in this paper. For the post and comment examples, we paraphrase them. }
    \label{fig:semantics}
\end{figure*}

We rely on a multi-task learning framework to predict 8 semantic categories (e.g., insightful, emotional support). This model was built on a combination of training data from human labels, user actions, and regex labels. We describe the semantic categories in Figure~\ref{fig:semantics}. For this analysis, we rely on a version of the model that relies only on comment text as input.

\subsection{Part I}
\subsubsection{Likelihood Ratio Test - Full and Social Context Models Compared to Baseline}
In this section, we show the Likelihood Ratio Test results comparing different model specifications to a baseline model that includes only post topics in Table~\ref{tab:lrt1}. We compare individual social context features (+Age, +Gender, +Friend Type) to the baseline model and find individual features lead to a better fit. When we compare to a Full model, which includes all social context signals, we see further gains. 

\begin{table*}[ht]
\resizebox{\textwidth}{!}{%
\begin{tabular}{l|llllllll}
\hline\hline
                                   & Surprised  & Funny & Insult & Emo  Support & Self-Disclosure & Worried & Thankful & Relaxed \\ \hline
Baseline     & \minus 735416    &  \minus 1142279     &  \minus 1121999      &    \minus 1021913                 &    \minus 789679             &   \minus 898347      &   \minus 1183726       & 543822        \\ 
+Age         & \minus 734805$^{***}$ &  \minus 1130343$^{***}$     &  \minus 1116308$^{***}$      &    \minus 1017417$^{***}$                 &      \minus 786774$^{***}$           &   \minus 895814$^{***}$      &  \minus 1176533$^{***}$        &   546275$^{***}$      \\ 
+Gender      & \minus 732373$^{***}$ &  \minus 1141036$^{***}$     & \minus 1115312$^{***}$       &   \minus 1014161$^{***}$                  &    \minus 789117$^{***}$             &  \minus 897948$^{***}$      &  \minus 1177248$^{***}$        &  544535$^{***}$       \\ 
+Friend & \minus 735340$^{***}$ & \minus 1141821$^{***}$      &   \minus 1121958$^{***}$     &  \minus 1021860$^{***}$                   &    \minus 789478$^{***}$             &   \minus 898180$^{***}$      &  \minus 1183699$^{***}$        &   543910$^{***}$      \\ 
Full   & \minus 731698$^{***}$ &  \minus 1128197$^{***}$     &   \minus 1109017$^{***}$    &   \minus 1009230$^{***}$                  &    \minus 786108$^{***}$             &   \minus 895229$^{***}$      &  \minus 1169517$^{***}$        &   547143$^{***}$      \\ \hline\hline
\end{tabular}%
}
\caption{We include results for likelihood ratio tests across the different semantic tone categories when comparing to a baseline model. We find better model fit for social context signals individually and all social context signals (Full) when compared to a Baseline model. *$p<$0.1; **$p<$0.05; ***$p<$0.01}
\label{tab:lrt1}
\end{table*}

\begin{table*}[ht]
\resizebox{\textwidth}{!}{%
\begin{tabular}{l|llllllll}
\hline\hline
                                   & Surprised  & Funny & Insult & Emo  Support & Self-Disclosure & Worried & Thankful & Relaxed \\ \hline
Age     & \minus 734805   &  \minus 1130343    &  \minus 1116308     &  \minus 1017417                   &  \minus 786774              &      \minus 895814   &   \minus 1176533       &  546275     \\ 
Gender         & \minus 732373 &  \minus 1141036    &  \minus 1115312    &    \minus 1014161                 & \minus 789117             & \minus 897948      &  \minus 1177248       &  544535      \\ 
Friend      & \minus 735340 &  \minus 1141821   &  \minus 1121958   &   \minus 1021860      &  \minus 789478           & \minus 898180      &  \minus 1183699    &   543910   \\ 
+Full & \minus 731698$^{***}$ &  \minus 1128197$^{***}$     &  \minus 1109017$^{***}$     &  \minus 1009230$^{***}$                  & \minus 786108$^{***}$               &   \minus 895229$^{***}$   &   \minus 1169517${***}$      &   547143$^{***}$     \\ \hline\hline

\end{tabular}%
}
\caption{We include results for likelihood ratio tests across the different semantic tone categories when comparing to a full model. The comparison is done by comparing the Full model to each social context signal individually. We see that the Full model provides better fit than a model with just a single social context signal. *$p<$0.1; **$p<$0.05; ***$p<$0.01.  }
\label{tab:lrt2} 
\end{table*}

\subsubsection{Likelihood Ratio Test - Full Compared to Individual Social Context Features}
We next do a Likelihood Ratio test where we compare the Full model to a model that has only one social context (Age, Gender, or Friend) signal. Across all semantic categories, we see in Table~\ref{tab:lrt2} that the Full model outperforms a model with just a single social context signal.


\subsection{Part II}

\subsubsection{LLM prompts}
In this section, we describe how we prompt the LLM to generate comments. Each prompt consists of three parts: background, context, and generation instructions. The background part provides a general description of the task and the user post, which is shared across all prompts:
\begin{quote}
\textit{You are a Facebook user and want to comment on a post from another Facebook user (the poster) in the most proper way based on the social relationship between you and the poster.} 

\textit{Here is the post made by the poster on Facebook: 
{[}Post{]}*\_\_ \{post\_text\} \_\_*.}
\end{quote}
The instruction part is as follows:
\begin{quote}
\textit{Generate a short comment right below the post, with ``{[}Comment{]}:'' prefixing the comment, in one paragraph. No explanation needed.}
\end{quote}
The context component varies between prompt types and provides different levels of social context information. Table \ref{prompts} presents the context part of different prompt types. The field \{friendship\_status\} can be one of the following:
\begin{itemize}
    \item close friends: ``\textit{you and the poster are very close friends and interact with each other frequently}''
    \item non-close friends: ``\textit{you and the poster are friends but not interact with each other much}''
    \item strangers: ``\textit{you and the poster are strangers}''
\end{itemize}

\begin{table*}[ht]
\centering
\begin{tabular}{p{2.5cm}|p{14cm}}
\hline
\hline
\textbf{Prompt type}    & \textbf{Context}  \\
\hline
\multirow{2}{*}{llm-all-context} &  Remember that \{friendship\_status\}.
Also remember that you are a \{commenter\_gender\} of age \\ & \{commenter\_age\} and the poster is a \{poster\_gender\} of age \{poster\_age\} \\
\hline 
llm-age-only & Remember that you are of age \{commenter\_age\} and the poster is of age \{poster\_age\} \\
\hline
llm-gender-only & Remember that you are a \{commenter\_gender\} and the poster is a \{poster\_gender\}. \\
\hline
llm-friends-only & Remember that \{friendship\_status\}. \\
\hline
\hline
\end{tabular}
\caption{Social context provided in different prompts.}
\label{prompts}
\end{table*}

\begin{table*}[hb]
\centering
\begin{tabular}{l|cccccccc}
\hline\hline
\textbf{\begin{tabular}[c]{@{}l@{}}Social context\\ factor\end{tabular}} & \textbf{Surprised}               & \textbf{Funny}                   & \textbf{Insult}                  & \textbf{\begin{tabular}[c]{@{}l@{}}Emotional\\ support\end{tabular}} & \textbf{\begin{tabular}[c]{@{}l@{}}Self\\ disclosure\end{tabular}} & \textbf{Worried}                 & \textbf{Thankful}                & \textbf{Inclusion}               \\ \hline
commenter\_male                                                          & \minus 0.046                           & 0.069                            & \minus 0.023                           & \minus 0.001                                                             & \cellcolor[HTML]{9cff8e}\minus 0.081$^{**}$                                   & \cellcolor[HTML]{F4CCCC}0.157$^{***}$ & 0.006                            & \cellcolor[HTML]{F4CCCC}0.069$^{**}$  \\
poster\_male                                                             & \minus 0.004                           & 0.005                            & \minus 0.06                            & \minus 0.026                                                             & 0.023                                                              & 0.011                            & \minus 0.004                           & \minus 0.017                           \\
\begin{tabular}[c]{@{}l@{}}commenter\_male $\times$\\ poster\_male\end{tabular} & 0.020                             & \cellcolor[HTML]{F4CCCC}0.263$^{***}$ & \cellcolor[HTML]{F4CCCC}0.157$^{**}$  & 0.050                                                               & \cellcolor[HTML]{9cff8e}\minus 0.080$^{*}$                                     & 0.022                            & \cellcolor[HTML]{F4CCCC}0.234$^{***}$ & 0.022                            \\ \hline
commenter\_age                                                           & \cellcolor[HTML]{F4CCCC}0.019$^{*}$   & \cellcolor[HTML]{9cff8e}\minus 0.056$^{**}$ & \minus 0.005                           & \cellcolor[HTML]{9cff8e}\minus 0.019$^{*}$                                    & 0.016                                                              & 0.008                            & 0.007                            & 0.002                            \\
poster\_age                                                              & \cellcolor[HTML]{F4CCCC}0.024$^{*}$   & 0.022                            & \minus 0.031                           & 0.003                                                              & \minus 0.013                                                             & \minus 0.001                           & \minus 0.016                           & \minus 0.001                           \\
\begin{tabular}[c]{@{}l@{}}commenter\_age $\times$\\ poster\_age\end{tabular}   & 0.013                            & \minus 0.01                            & 0.010                             & \minus 0.009                                                             & 0                                                                  & 0.007                            & \minus 0.01                            & \minus 0.005                           \\ \hline
non-close friends                                                        & \cellcolor[HTML]{F4CCCC}0.083$^{***}$ & \cellcolor[HTML]{F4CCCC}0.395$^{***}$ & \cellcolor[HTML]{F4CCCC}0.255$^{***}$ & 0.011                                                              & \cellcolor[HTML]{F4CCCC}0.283$^{***}$                                   & \minus 0.009                           & \cellcolor[HTML]{F4CCCC}0.202$^{***}$ & \cellcolor[HTML]{F4CCCC}0.097$^{***}$ \\
strangers                                                                & \cellcolor[HTML]{F4CCCC}0.104$^{***}$ & \cellcolor[HTML]{F4CCCC}0.318$^{***}$ & \cellcolor[HTML]{F4CCCC}0.13$^{*}$    & \cellcolor[HTML]{F4CCCC}0.072$^{*}$                                     & \cellcolor[HTML]{F4CCCC}0.238$^{***}$                                   & \minus 0.021                           & 0.066                            & \cellcolor[HTML]{F4CCCC}0.094$^{**}$  \\ \hline \hline
\end{tabular}
\caption{The impact of providing context on the discrepancy between generated comments and human comments. The discrepancy is calculated as the difference in fitting coefficients between the regression models for generated comments and that for human comments. Significant changes are highlighted: red indicates that adding context increased the discrepancy, while green shows a reduction in discrepancy due to context. * $p<0.05$, ** $p<0.01$, *** $p<0.001$. p-values were obtained via two-sample t-test.}
\label{appendix-table-8}
\end{table*}

\subsubsection{Comparison between Approach 1 and Approach 2}
\label{sec:compare}
This section utilizes a straightforward mathematical model to contrast Approach 1 and Approach 2 from Part II. We demonstrate that Approach 1 considers both between-group and within-group differences, whereas Approach 2 focuses solely on between-group differences.

We frame our analysis based on the social context predictor $X$, with two response variables: $Y$ representing the tone of human comments, and $Z$ indicating the tone of generated comments. Each approach employs distinct metrics to assess the similarity between $Y$ and $Z$. For simplicity, we consider a binary predictor: $X_i=1$ indicates group 1 and $X_i=0$ indicates group 0. 

In Approach 1, we employ the regression model
\[
Z_i = a_{zy} + b_{zy} Y_i + e_i,
\]
where the coefficient $b_{zy}$ is
\begin{equation}
b_{zy} = \frac{\mathrm{Cov}(Y, Z)}{\mathrm{Var}(Y)}.
\label{b_zy}
\end{equation}
Here, $\mathrm{Cov}(Y,Z)$ represents the covariance between $Y$ and $Z$, and $\mathrm{Var}(Y)$ signifies the variance of $Y$. These statistics are partitioned into three components: (co)variance from group 0 ($x=0$), (co)variance from group 1 ($x=1$), and between-group (co)variance, calculated as follows:
\begin{equation*}
\begin{split}
\mathrm{Cov}(Y, Z) =& \frac{n_0}{n}\mathrm{Cov}(Y, Z)_{x=0} + \frac{n_1}{n}\mathrm{Cov}(Y, Z)_{x=1} \\ 
& + \frac{n_0n_1}{n^2}{} (\bar Z_1 - \bar Z_0)(\bar Y_1 - \bar Y_0),
\end{split}
\end{equation*}
\begin{equation*}
\mathrm{Var}(Y) = \frac{n_0}{n} \mathrm{Var}(Y)_{x=0} + \frac{n_1}{n} \mathrm{Var}(Y)_{x=1} + \frac{n_0n_1}{n^2}(\bar Y_1 - \bar Y_0)^2.
\end{equation*}
$\bar Y_0$ and $\bar Y_1$ denotes the mean values of $Y$ within group 0 and 1, respectively. Similarly, $\bar Z_0$ and $\bar Z_1$ are the mean values of $Z$ in these groups. $n_0$ and $n_1$ are sample size of groups 0 and 1, respectively, with $n=n_0 + n_1$ representing the total sample size. 

Therefore, both the within-group and between-group (co)variance influence the regression coefficients $b_{zy}$. If the between-group (co)variance is significant compared to within-group (co)variance, $b_{zy}$ approximates the ratio of between-group differences, $(\bar Z_1 - \bar Z_0)/(\bar Y_1 -\bar Y_0)$. Conversely, if the within-group (co)variance is dominant, $b_{zy}$ approximates $(n_0\mathrm{Cov}(Y, Z)_{x=0} + n_1\mathrm{Cov}(Y, Z)_{x=1} )/(n\mathrm{Var}(Y))$.

For approach 2, we estimate two separate regressions,
\begin{equation*}
\begin{split}
Y_i = a_y + b_y X_i + e_{yi},\\
Z_i = a_z + b_z X_i + e_{zi}.
\end{split}
\end{equation*}
We then calculate the difference between regression coefficients,
\[
\delta_{zy} = |b_z - b_y|.
\label{delta_zy}
\]
This calculation is straightforward for a single variable predictor, resulting in
\begin{equation}
\delta_{zy} = |\bar Z_1 - \bar Z_0 - (\bar Y_1 - \bar Y_0)|.
\end{equation}

Comparing this with the coefficient $b_{zy}$ from Approach 1, $\delta_{zy}$ is solely dependent on between-group differences. In contrast, $b_{zy}$ incorporates both within-group differences and within-group variability, facilitating individual-level comparisons.


\begin{figure*}[h]
    \centering
    \includegraphics[width=0.8\textwidth]{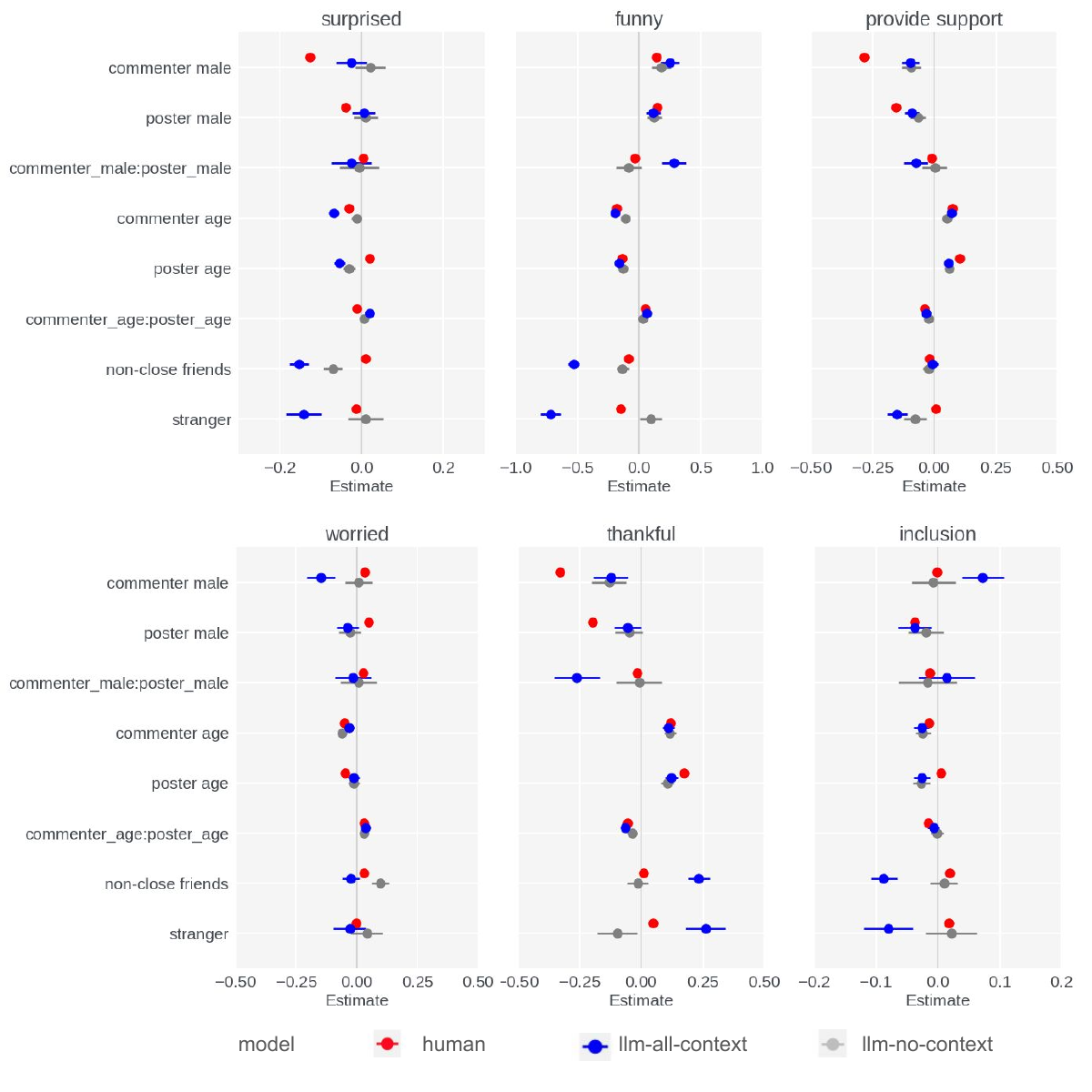}
    \caption{Regression coefficients of social context factors in the fitted models for remaining scores, including surprised, funny, provide support, worried, thankful, and inclusion.}
    \label{fig-3-appendix}
\end{figure*}

\subsubsection{Prompt engineering}
To test whether our conclusion relies on how the social context information is presented in the prompt, in this section we study how variation of the prompt affects the discrepancy between generated comments and human comments. We tested the following ideas,
\begin{itemize}
    \item Adjusting the order of social context information in the prompt. More specifically, we switched the order of age, gender, and friendship type information in the context part of the prompts, as demonstrated in Table \ref{prompts}.
    \item Zero-shot chain of thoughts (CoT). 
    Following the idea of zero-shot-CoT, we added the following sentence to the instruction part of the prompt: ``\textit{Think step by step with explanations}''.
    \item Rephrasing the social context in the prompt. Our results imply that the generated response is very sensitive to the friendship type presented in the social context. To test the reliability, we rephrase the friendship type from ``strangers'' to ``non-friends''.
\end{itemize}
The results are summarized in \textcolor{black}{Table~\ref{prompt_eng}}, which shows while some prompt engineering techniques helped to reduce the discrepancy, they did not alter the main conclusion of our findings.

As illustrated in \textcolor{black}{Table~\ref{prompt_eng}}, altering the order of social context information in the prompt significantly reduces discrepancies. This finding aligns with previous studies indicating that the sequence of instructions in prompts affects LLM performance. Additionally, employing CoT reasoning also decreases discrepancies. However, modifying the description of friendship types has a marginal effect. Despite these adjustments, the no-context prompt still exhibits the lowest discrepancy, underscoring the LLM's limited ability to respond appropriately to social context.

\begin{table}[H]
\centering
\begin{tabular}[width=\linewidth]{l|c}
\hhline{==}
\textbf{Prompt}                & \multicolumn{1}{l}{\textbf{Average discrepancy}} \\ 
\hline
baseline all\_context & 0.803             \\
\hline
gender\_age\_friends  & \minus 0.133                                  \\
age\_gender\_friends  & \minus 0.084                                 \\
zero\_shot\_cot       & \minus 0.063                                 \\ 
non\_friends & +0.010\\
\hline\hline 
\end{tabular}
\caption{Tuning the prompt helps to reduce but cannot eliminate the discrepancy $\bar{\hat{\delta}}$. First row shows the average discrepancy from the baseline all context prompt, followed by relative changes after modifying the prompt.}
\label{prompt_eng}
\end{table}

\end{document}